\documentclass[letterpaper,10pt,conference]{IEEEtran}

\usepackage[english]{babel}

\usepackage{amssymb}
\usepackage{amsmath}
\usepackage{hyperref}
\usepackage{cleveref}
\usepackage{graphicx}
\usepackage{latexsym}
\usepackage{algorithm}
\usepackage{algpseudocode}
\usepackage{todonotes}
\usepackage{newclude}
\usepackage{multicol}
\usepackage{tabularx}
\usepackage{booktabs}
\usepackage{threeparttable}

\usepackage{enumitem}
\usepackage{listings}
\usepackage{combelow}
\usepackage{subfigure}
\usepackage{bigfoot}

\usepackage{url}
\newcommand\rurl[1]{%
  \href{http://#1}{\nolinkurl{#1}}%
}

\usepackage[binary-units]{siunitx}
\usepackage[acronym]{glossaries}
\newacronym{fmcw}{FMCW}{Frequency-Modulated Continuous-Wave}
\newacronym{cnn}{CNN}{Convolutional Neural Network}
\newacronym{vlad}{VLAD}{Vector of Locally Aggregated Descriptors}
\newacronym{dl}{DL}{Deep Learning}
\newacronym{wsl}{WSL}{Weakly-Supervised Learning}
\newacronym{vo}{VO}{Visual Odometry}
\newacronym{gps}{GPS}{Global Positioning System}
\newacronym{ro}{RO}{Radar Odometry}
\newacronym{uwb}{UWB}{Ultra Wide Band}
\newacronym{fft}{FFT}{Fast Fourier Transform}
\newacronym{slam}{SLAM}{Simultaneous Localisation and Mapping}
\newacronym{mmw}{MMW}{Millimetre-Wave}
\newacronym{fcnn}{FCNN}{Fully Convolutional Neural Network}
\newacronym{lidar}{LiDAR}{Light Detection and Ranging}
\newacronym{nn}{NN}{nearest neighbour}
\newacronym{auc}{AUC}{Area-under-Curve}
\newacronym{fov}{FOV}{field-of-view}
\newacronym{pr}{PR}{precision and recall}

\glsdisablehyper

\definecolor{CommentStef}{rgb}{0.2,0.8,0.2}

\definecolor{CommentMatt}{rgb}{1,0.2,0}

\definecolor{CommentDani}{rgb}{0,0,1}

\definecolor{CommentPaul}{rgb}{0.9,0,0}

\definecolor{CommentReview}{rgb}{0.9,0.6,0.2}

\newcommand{\vanilla}{{\sc vgg-16/netvlad}}
\newcommand{\radvlad}{{\sc ours}}

\IEEEoverridecommandlockouts
\crefname{table}{Table}{Tables}
\crefname{figure}{Figure}{Figures}
\crefname{section}{Section}{Sections}

\title{Kidnapped Radar: Topological Radar Localisation using Rotationally-Invariant Metric Learning}
\author{\cb{S}tefan S\u{a}ftescu, Matthew Gadd, Daniele De Martini, Dan Barnes, and Paul Newman\\
Oxford Robotics Institute, Dept. Engineering Science, University of Oxford, UK.\\\texttt{\{stefan,mattgadd,daniele,dbarnes,pnewman\}@robots.ox.ac.uk}}

\begin{document}

\maketitle

\begin{abstract}
This paper presents a system for robust, large-scale topological localisation using \gls{fmcw} scanning radar.
We learn a metric space for embedding polar radar scans using \acrshort{cnn} and NetVLAD architectures traditionally applied to the visual domain.
However, we tailor the feature extraction for more suitability to the polar nature of radar scan formation using cylindrical convolutions, anti-aliasing blurring, and azimuth-wise max-pooling; all in order to bolster the rotational invariance.
The enforced metric space is then used to encode a reference trajectory, serving as a map, which is queried for \glspl{nn} for recognition of places at run-time.
We demonstrate the performance of our topological localisation system over the course of many repeat forays using the largest radar-focused mobile autonomy dataset released to date, totalling \SI{280}{\kilo\metre} of urban driving, a small portion of which we also use to learn the weights of the modified architecture.
As this work represents a novel application for \gls{fmcw} radar, we analyse the utility of the proposed method via a comprehensive set of metrics which provide insight into the efficacy when used in a realistic system, showing improved performance over the root architecture even in the face of random rotational perturbation.
\end{abstract}

\glsresetall

\section{Introduction}%
\label{sec:introduction}

\begin{figure}[h!]
    \centering
    \includegraphics[width=0.8\columnwidth]{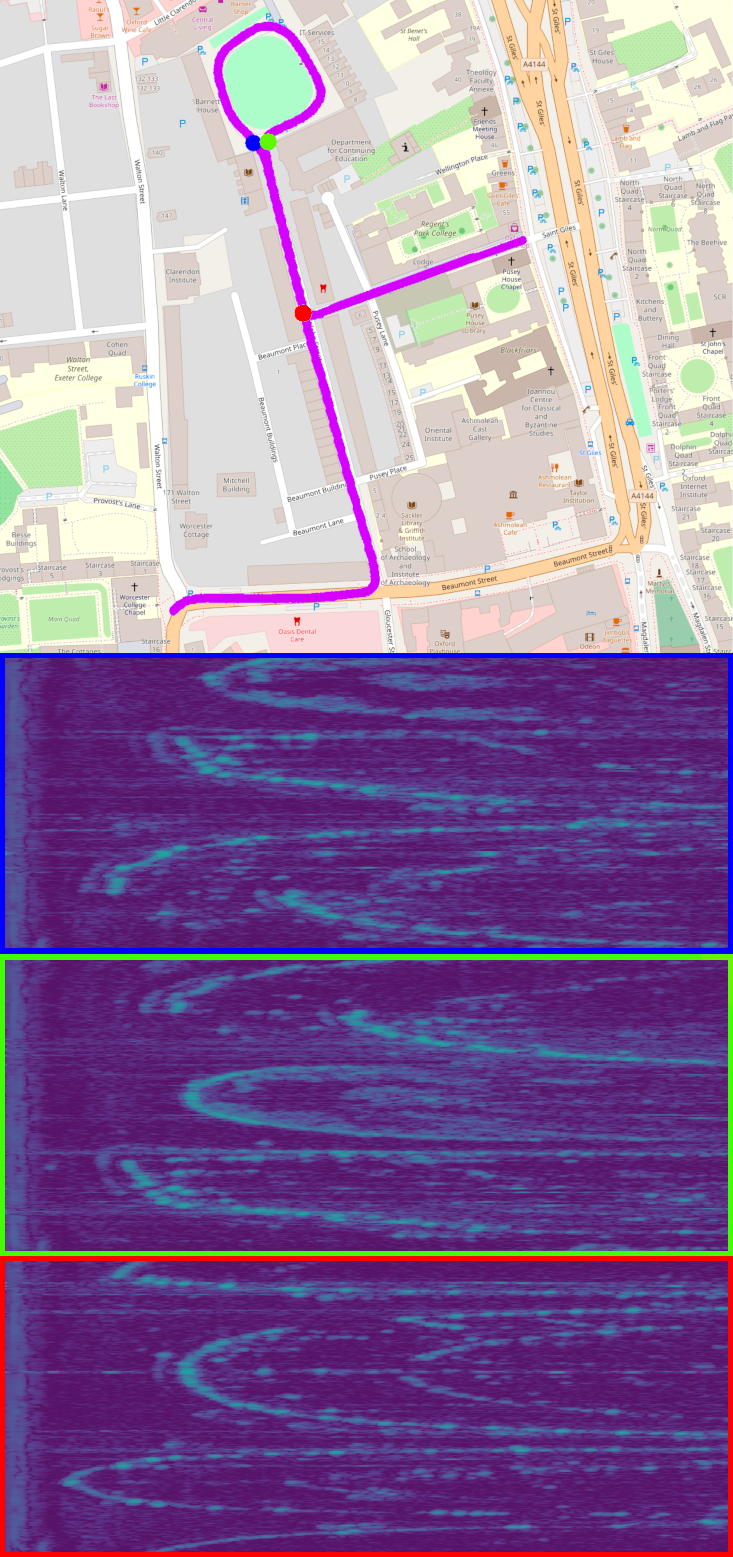}
    \caption{Place recognition using \gls{fmcw} radar: given an online query radar scan (blue dot on map and blue-framed radar image), the aim is to retrieve a correct match (green), disregarding the incorrect, although similar, radar scan the map also represents (red) and despite the obvious rotational offset.
    Map data copyrighted OpenStreetMap~\cite{OpenStreetMap,haklay2008openstreetmap} contributors and available from \rurl{openstreetmap.org}.}
    \label{fig:osm_se2}
    \vspace{-0.3cm}
\end{figure}

In order for autonomous vehicles to travel safely at higher speeds or operate in wide-open spaces where there is a dearth of distinct features, a new level of robust sensing is required.
\gls{fmcw} radar satisfies these requirements, thriving in all environmental conditions (rain, snow, dust, fog, or direct sunlight), providing a \SI{360}{\degree} view of the scene, and detecting targets at ranges of up to hundreds of metres with centimetre-scale precision.

Indeed, it has been shown that this class of radar can be effectively used for accurate motion estimation in various challenging environments using scan matching and data association of hand-crafted features~\cite{2018ICRA_cen,2019ICRA_cen,2019ITSC_aldera}.
Real-time deployment of this type of approach is possible by preprocessing the radar measurement stream and easing the data association burden~\cite{2019ICRA_aldera}.
The present state-of-the-art in \gls{ro} learns masks to apply to the radar data stream as well as an artefact- and distraction-free embedding in an end-to-end fashion~\cite{barnes2019masking}.

\begin{figure}
    \centering
    \includegraphics[width=0.9\columnwidth]{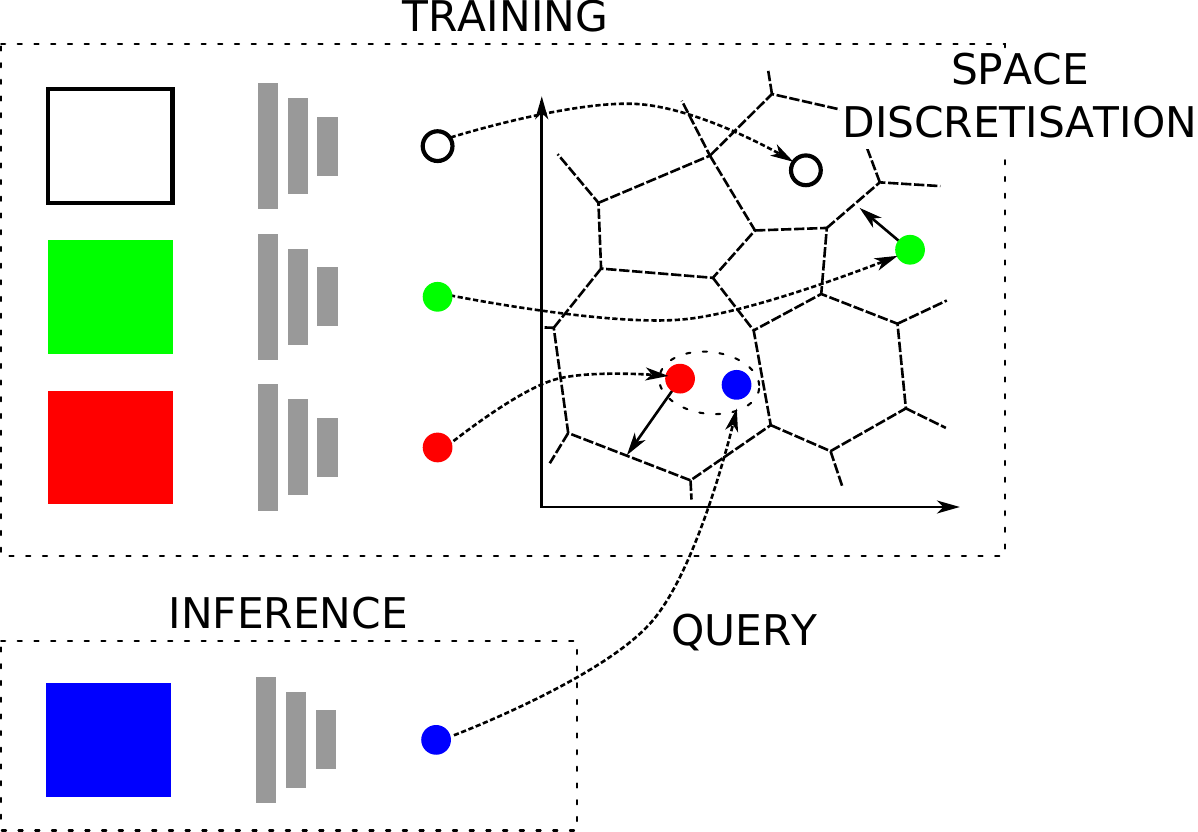}
    \caption{The \gls{fmcw} radar place recognition pipeline.
    The offline stages of the pipeline involve \emph{enforcing} and \emph{discretising} the metric space, while the online stages involve \emph{inference} to represent the place the robot currently finds itself within in terms of the learned knowledge and \emph{querying} the discretised space, in this case depicted using a Voronoi-like structure, which encodes the trajectory of the robot.}
    \label{fig:pipeline}
    \vspace{-0.3cm}
\end{figure}

With these modern capabilities, it is currently possible to apply \gls{fmcw} radar to the construction of accurate map representations for use in an autonomy stack.
Metric pose estimation in an unconstrained search over all places represented in the map is therefore feasible but would not scale well with the size of the environment.
In the best case, when using heuristics for constraining the graph search, eventual drift in the ego-motion is likely to invalidate any reported poses. 

This paper thus presents a system which reproduces and advances in the radar domain the current capabilities in visual place recognition to produce topological localisation suggestions which we envision being used downstream for metric pose estimation.
We believe that this represents the first occasion in which place recognition is performed for the \gls{fmcw} class of radar.
As our radar produces \SI{360}{\degree} scans we note that, unlike narrow \gls{fov} cameras, the orientation of the sensor is irrelevant for place recognition: whether the vehicle is facing east or west on a street, the scan will be the same up to rotation.
With this observation in mind, we design a \gls{fcnn} which is quasi-invariant to rotations of the input scans, and learn an embedding space which we can query for similarity between a reference trajectory and the live scan.

This paper proceeds by reviewing existing literature in~\cref{sec:related}, followed by a brief preliminary discussion of radar image formation in~\cref{sec:prelim}.
\cref{sec:system} gives an overview of our system and motivates the desired online operation, followed by a description in \cref{sec:network} of an offline learning stage which satisfies these design principles.
Finally,~\cref{sec:experimental} presents necessary details for implementation as well as our experimental philosophy before~\cref{sec:results}, where we report our results.

\section{Related Work}%
\label{sec:related}

Place recognition is a somewhat consolidated procedure in the camera sensor modality.
A brief history of the community's progress in this regard includes: probabilistic models around bag-of-words image representations~\cite{cummins2008fab}, sequence-based approaches~\cite{milford2012seqslam}, and more recently by extracting features from the responses of \gls{cnn} layers and subsequent use of these features for image comparison~\cite{khaliq2018holistic}.

There has also been extensive investigation of \gls{lidar}-based place recognition, often relying on geometrical features to overcome extreme appearance change, including systems based on: matching of 3D segments~\cite{dube2017segmatch}, semantic graph descriptor matching~\cite{gawel2018x}, learned discriminative global features~\cite{angelina2018pointnetvlad}, and combining the benefits of geometry and appearance by coupling the conventional geometric information from the \gls{lidar} with its calibrated intensity return~\cite{guo2019local}.

Besides its superior range and despite its lower spatial resolution, \gls{mmw} radar often overcomes the shortcomings of laser, monocular, or stereo vision because it can operate successfully in dust, fog, blizzards, and poorly lit scenes~\cite{reina2015radar}.
In~\cite{mielle2019comparative} it is shown in the context of a \gls{slam} system that while producing slightly less accurate maps than \glspl{lidar}, radars are capable of capturing details such as corners and small walls.

Place recognition with \gls{uwb} radar is presented in~\cite{takeuchi2015localization} by matching received signals to a database of waveforms which represent signatures of places.
Although the \gls{uwb} class of radar is capable of very high update rates, it is shown in~\cite{figueroa2016comparison} that the \gls{fmcw} class is superior in raw measurement quality, measured maximum range, and worst-case precision.
As our system is eventually to be included in a larger framework which must yield precise pose estimation (c.f.~\cref{sec:introduction}), the \gls{fmcw} class is therefore our preferred sensor.
Furthermore, evaluation in~\cite{takeuchi2015localization} was performed in indoor and forested environments, whereas our work is deployed in built environments representative of urban driving.

\section{Preliminaries}
\label{sec:prelim}

We use a \gls{fmcw} scanning radar which rotates about its vertical axis while sensing the environment continuously through the transmission and reception of frequency-modulated radio waves.
While rotating, the sensor inspects one angular portion (\textit{azimuth}) of space at a time and receives a power signal that is a function of reflectivity, size, and orientation of objects at that specific azimuth and at a particular distance.
The radar takes measurements along an azimuth at one of a number of discrete intervals and returns a list of power readings, or \textit{bins}.
We call one full rotation across all azimuths a \textit{scan}, some examples of which are shown in~\cref{fig:osm_se2,fig:batch}.

\section{System Overview}%
\label{sec:system}

\cref{fig:pipeline} depicts the overall method.
As motivated in~\cref{sec:introduction}, we require a system which produces topological localisation suggestions, used downstream for metric pose estimation.

\subsection{Design requirements}

To satisfy our design outcomes, we do not require a mature \gls{slam} system which models environment concurrently with estimating the state of the sensor~\cite{cadena2016past}.
Instead, we approach the place recognition problem as a \gls{nn} search in a multidimensional space, where each portion of the space describes a different place and points within a portion represent different views of a place.

We consider this approach well-posed as the invariance of radar measurements to changing environmental conditions (such as illumination, rain, fog, etc) implies that a map built from a single experience of the route will likely be of good utility over the course of several months or seasons, as only changes to the structure of the scene itself (e.g. building construction) will present significant variation in scene appearance.

\subsection{Offline learning}

To achieve our requirements, good metric embeddings of the polar radar measurements are required which can be used to create a map of the environment in which the robot will operate by encoding places of interest offline.

Generation of these embeddings is delegated to an encoder network (c.f.~\cref{sec:network}) which extracts information from the radar measurements and compresses them within the multidimensional space.
The \textit{training} procedure enforces that the network will learn this transformation.

Because of the geographical scale of the environment which must be encoded for representation (large urban centres), exploitation of common data structure techniques to discretise this space for fast lookups is essential to reduce the \gls{nn} search complexity.
In the results discussed in \cref{sec:results}, a k-dimensional tree~\cite{bentley1975multidimensional} structure is used.
This choice guarantees the exactness of the search result, thus not influencing our discussion of the accuracy of the learned representation.
Other CPU- or GPU-based methods allow for faster, although approximate, searches~\cite{malkov16,wieschollek_2016_CVPR}.

\subsection{Mapping and Localisation}

At deployment time, inference on the network involves a feed-forward pass of a single radar scan, resulting in an embedding, i.e.~a point in a multidimensional space.
The metric nature of the learned space allows us to obtain a measure of similarity to the stored database and query it for topological localisation candidates.
To this end, the discretised space discussed above is traversed for the closest places from the database of known locations within a certain embedding distance threshold.
Alternatively, a fixed-size set, $N$, of top scoring candidates are all considered.

\begin{figure}
    \centering
    \includegraphics[width=0.75\columnwidth]{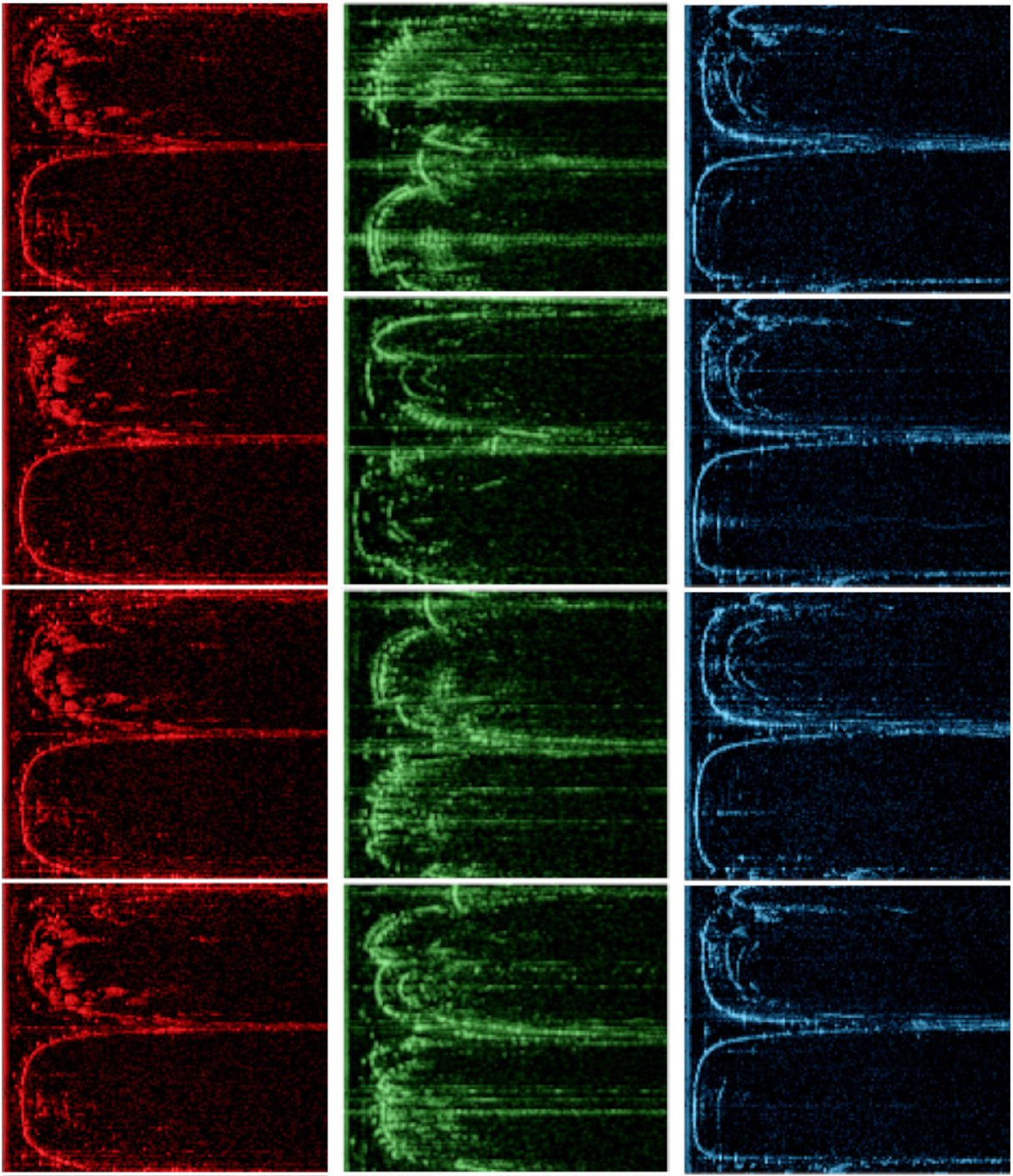}
    \caption{A (contrast-enhanced) visualisation of a batch of radar scans input to our network during training.
    Each scan shown is a range-versus-azimuth grayscale image.
    Batches are constructed such that there is no overlap in the radar sensing horizon between anchor scans (top row).
    Scans with returns painted with the same colour as an anchor (red, green, or blue) are marked as positive examples of topological localisations (columns).}
    \label{fig:batch}
    \vspace{-0.3cm}
\end{figure}

\section{Learning the Metric Space}%
\label{sec:network}

To learn filters and cluster centres which help distinguish polar radar images for place recognition we use NetVLAD~\cite{arandjelovic2016netvlad} with VGG-16~\cite{simonyan2014very} as a front-end feature extractor.
Specifically, we modify the implementation described in~\cite{cieslewski2018data}\footnote{\rurl{github.com/uzh-rpg/netvlad\_tf\_open}} to make the network invariant to the orientation of input radar scans.
We refer to the original architecture as \vanilla~and our proposed architecture as \radvlad.

\subsection{Feature extraction}

With similar motivation to~\cite{wang2018omnidirectional} we apply circular padding to the \gls{cnn} feature spaces to reflect the fact that the polar representation of the assembled \gls{fft} returns has no true image boundary along the azimuth axis.
This provides rotation \emph{equivariance} throughout the network.

A common design in \gls{cnn}s is to downsample feature maps every few layers to reduce computation cost and increase the input area a single network filter receives information from.
As noted by~\cite{zhang2019making}, this breaks the translation equivariance \gls{cnn}s are usually assumed to have and therefore also the rotational equivariance provided by circular padding.
We apply the same solution from~\cite{zhang2019making} in our network by replacing the usual max-pooling with stride 2 used for downsampling in VGG-16 with stride 1 max-pooling, followed by a stride 2 Gaussian blur.
While this does not fully restore rotational equivariance,~\cite{zhang2019making} show that it greatly reduces the aliasing introduced by downsampling.

Finally, to make the network rotationally \emph{invariant} (up to the small aliasing that remains from downsampling), we peform max-pooling upon the last feature map along the azimuth axis.
As the $\mathrm{max}$ function is commutative and associative, and the last feature map is rotationally equivariant, the result will be rotationally invariant.

\subsection{Enforcing the metric space}

To enforce the metric space, we perform online triplet mining and apply the triplet loss described in~\cite{schroff2015facenet}.
Loop closure labels are taken from a ground truth dataset, which will be discussed in~\cref{sec:experimental}.
Batches, as illustrated in~\cref{fig:batch}, are constructed such that there is no overlap of the radar sensing horizon between a candidate radar scan and any anchor scan already sampled for the batch.

\subsection{Training details and hyperparameters}

Due to memory limitations on our graphical compute hardware, we crop the last \num{178} range bins and scale the width by a factor of \num{8} such that the original $400{\times}{3768}$ polar radar scans are input to the network with resolution $400{\times}{450}$ (c.f.~\cref{fig:batch}).
As the azimuth axis remains unscaled, this does not affect rotational invariance.

When finetuning either the original architecture or our proposed modified architecture, we initialise internal weights with the publicly available checkpoint {\small \verb|vd16_pitts30k_conv5_3_vlad_preL2_intra_white|}, corresponding to the best performing model described in~\cite{cieslewski2018data}.

Our learning rate schedule applies a linear decay function initialised at \SI{1e-4}{} and settling to \SI{5E-6}{} at \SI{5000}{} steps~\cite{bengio2012practical}.
We terminate learning at \num{500000} steps in all cases.
We use gradient clipping to limit the magnitude of the backpropagated gradients to a value of \SI{80}{}~\cite{goodfellow2016deep}.
An $L2$ vector norm is applied to regularise the weights with a scale of \SI{1E-7}{}.
We use two one-dimensional gaussian blur kernels with size \num{7} and standard deviation of \num{1}.

No augmentation has been performed on the training dataset.
In particular, we did not randomly rotate the input scans during training, in order to show the resilience of the rotationally invariant design of our network architecture, as assessed in~\cref{sec:results}.

\begin{figure}
    \centering
    \includegraphics[width=\columnwidth]{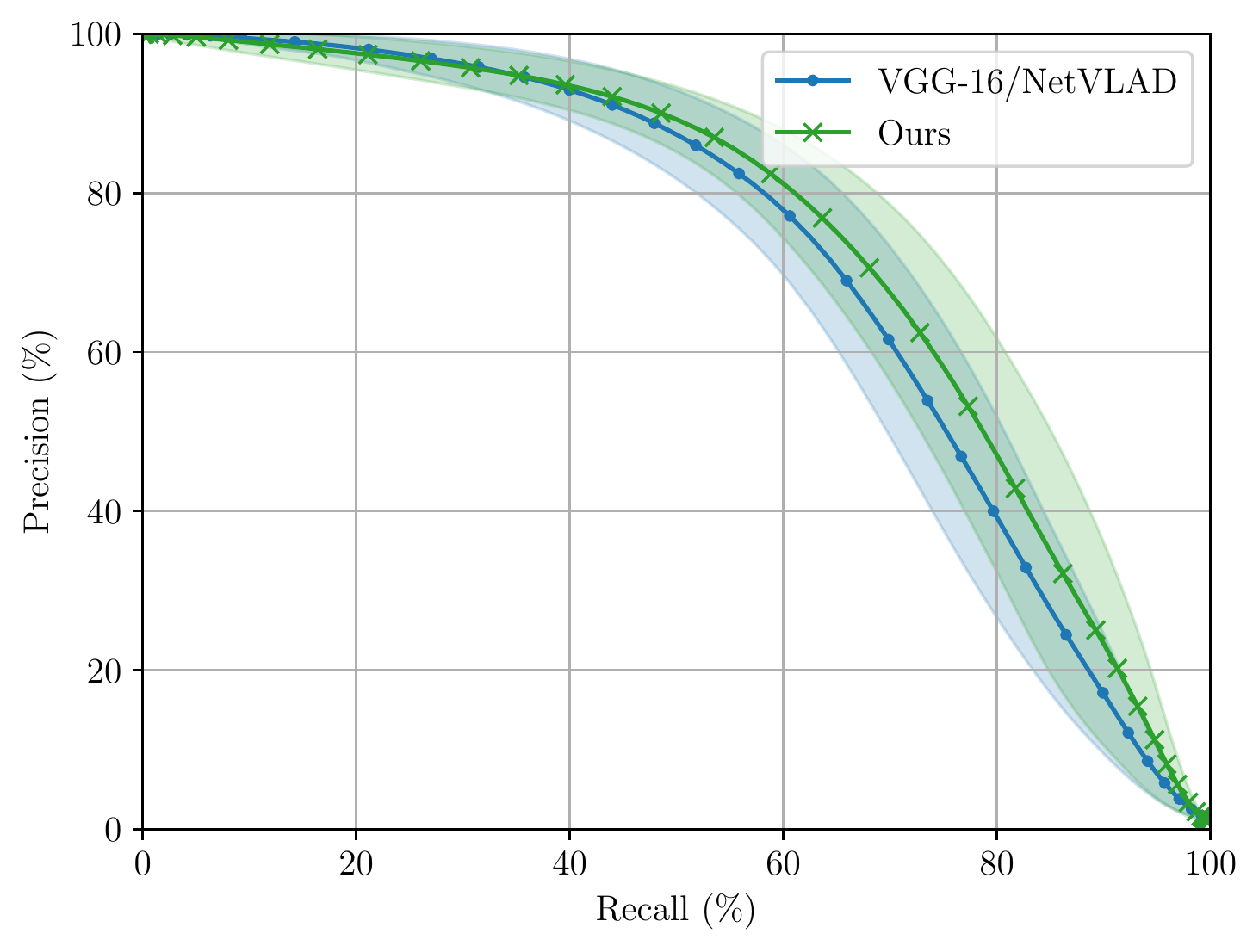}
    \caption{
    Average \gls{pr} curves and one standard-deviation bounds when using the learned representation to map an environment once and repeatedly localise consecutive trajectories against the static reference trajectory.
    The corresponding maximum $F_{1}$ scores are $0.70 \pm 0.04$ for \radvlad~(green) as compared to $0.68 \pm 0.04$ for \vanilla~(blue). 
    The corresponding maximum $F_{0.5}$ scores are $0.71 \pm 0.04$ for \radvlad~as compared to $0.69 \pm 0.04$ for \vanilla. 
    The corresponding maximum $F_{2}$ scores are $0.77 \pm 0.04$ for \radvlad~as compared to $0.76 \pm 0.04$ for \vanilla. 
    }
    \label{fig:rrcd_rr0_pr_curves}
    \vspace{-0.3cm}
\end{figure}

\section{Experimental Setup}%
\label{sec:experimental}

This section details our experimental setup and philosophy.
 
\subsection{Platform and sensor specifications}

The experiments are performed using data collected from the \textit{Oxford RobotCar} platform~\cite{RobotCarDatasetIJRR}.
The vehicle, as described in the recently released \textit{Oxford Radar RobotCar Dataset}~\cite{OxfordRadarRobotCarDataset}, is fitted with a CTS350-X Navtech \gls{fmcw} scanning radar without Doppler information, mounted on top of the platform with an axis of rotation perpendicular to the driving surface.
This radar is characterised by an operating frequency of \SIrange{76}{77}{\giga\hertz}, yielding  up to \SI{3768}{} range readings with a resolution of \SI{4.38}{\centi\meter} (a total range of \SI{165}{\metre}), each constituting one of the \SI{400}{} azimuth readings with a resolution of \SI{0.9}{\degree} and a scan rotation rate of \SI{4}{\hertz}.

\begin{figure}
    \centering
    \includegraphics[width=\columnwidth]{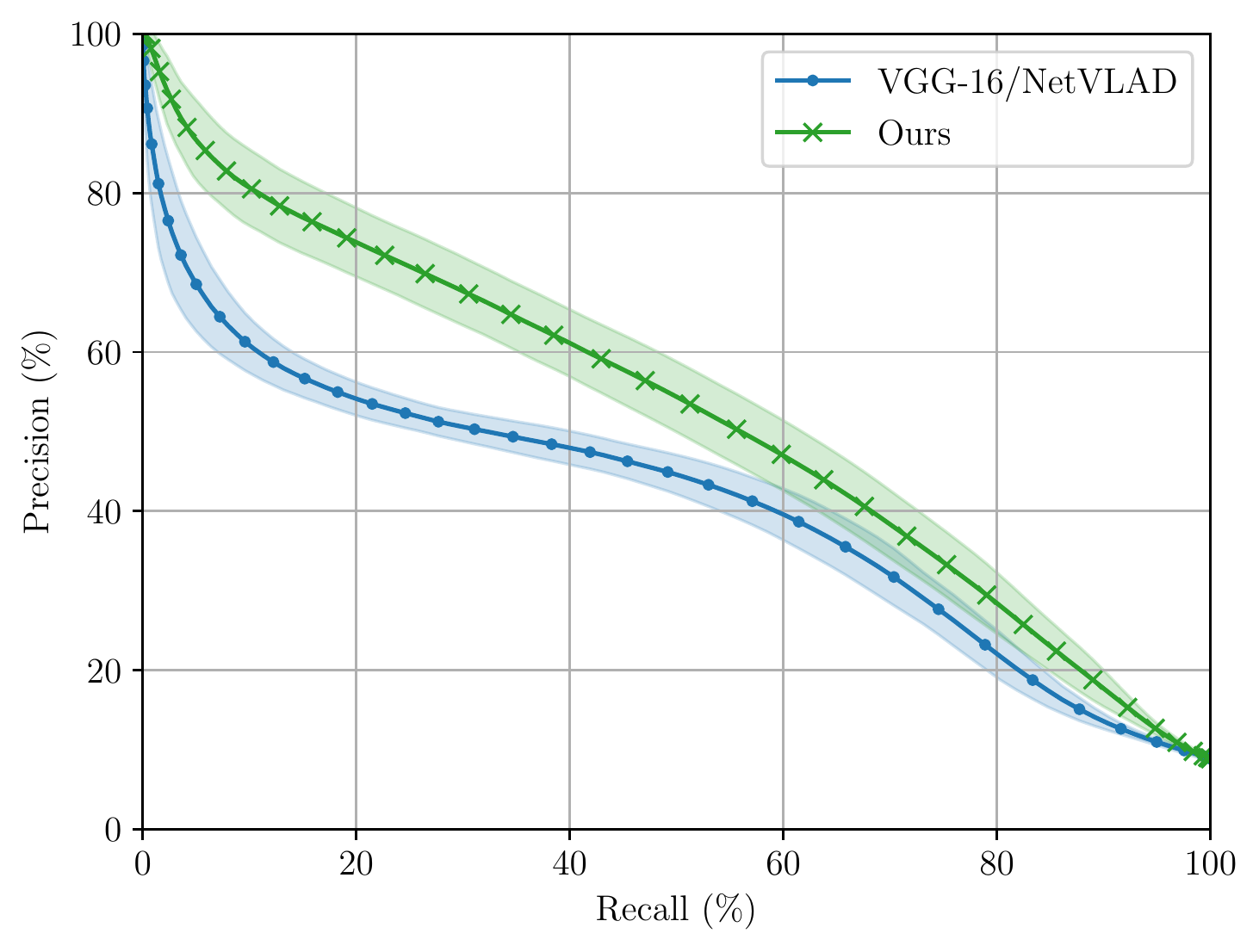}
    \caption{
    Average \gls{pr} curves and one standard-deviation bounds when using the learned representation to map a difficult, \emph{unseen} environment with backwards traversals, repeatedly localising consecutive trajectories in the \textit{test} split.
    The corresponding maximum $F_{1}$ scores are $0.53 \pm 0.03$ for \radvlad~(green) as compared to $0.48 \pm 0.02$ for \vanilla~(blue). 
    The corresponding maximum $F_{0.5}$ scores are $0.60 \pm 0.03$ for \radvlad~as compared to $0.57 \pm 0.02$ for \vanilla. 
    The corresponding maximum $F_{2}$ scores are $0.55 \pm 0.03$ for \radvlad~as compared to $0.46 \pm 0.02$ for \vanilla. 
    }
    \label{fig:test_rr0_pr_curves}
    \vspace{-0.3cm}
\end{figure}

\subsection{Ground truth location}
\label{sec:experimental:groundtruth}

For ground truth location, we manipulate the accompanying ground truth odometry described in~\cite{OxfordRadarRobotCarDataset}\footnote{\rurl{ori.ox.ac.uk/datasets/radar-robotcar-dataset}} which is computed by a global optimisation using \gls{gps}, robust \gls{vo}~\cite{barnes2018driven}, and visual loop closures from FAB-MAP~\cite{cummins2008fab}.

As each ground truth odometry file does not begin at the same location in the urban environment, we manually selected (for each of the \num{32} trajectories) a moment during which the vehicle was stationary at a common point and manually aligned the ground traces.
Furthermore, we align the ground traces manually by introducing a small rotational offset to account for differing attitudes at those instances.

The ground truth is preprocessed offline to capture the subsets of nodes that are at a maximum predefined distance, creating a graph-structured database that can easily be queried for triplets of nodes for training the system.

\subsection{Dataset demarcation}
\label{sec:experimental:demarc}

Each approximately \SI{9}{\kilo\metre} trajectory in the Oxford city centre was divided into three distinct portions: \textit{train}, \textit{valid}, and \textit{test}.
The maximum radar range was foreshortened due to the cluttered nature of the urban environment and we were thus able to specifically design the sets such that they did not overlap, padding the splits where necessary.

\cref{fig:osm_se2} shows the \gls{gps} trace of the \textit{test} split, which was specifically selected as the vehicle traverses a portion of the route in the opposite direction.
The \textit{valid} split selected was quite simple, consisting of two straight periods of driving separated by a right turn. 

\subsection{Trajectory pairs}

The network is trained with ground truth topological localisations between two reserved trajectories\footnote{{\small \verb|2019-01-10-11-46-21-radar-oxford-10k|} and\\ {\small \verb|2019-01-10-14-50-05-radar-oxford-10k|} from\\\rurl{ori.ox.ac.uk/datasets/radar-robotcar-dataset/datasets}}.

A large part of our analysis focuses on the principal scenario we propose, a typical teach-and-repeat session, in which all remaining trajectories in the dataset (excluding the partial traversals) are localised against a map\footnote{{\small{\verb|2019-01-10-12-32-52-radar-oxford-10k|}} from\\\rurl{ori.ox.ac.uk/datasets/radar-robotcar-dataset/datasets}} built from the first trajectory that we did not use to optimise the network weights, totalling \num{27} trajectory pairs with the same map but a different localisation run.

However, results on the test sets highlight the ability of the network to generalise: these are an indication of the performance of the system when deployed in environments which the network has not been trained on.

\begin{figure}
    \centering
    \includegraphics[width=\columnwidth]{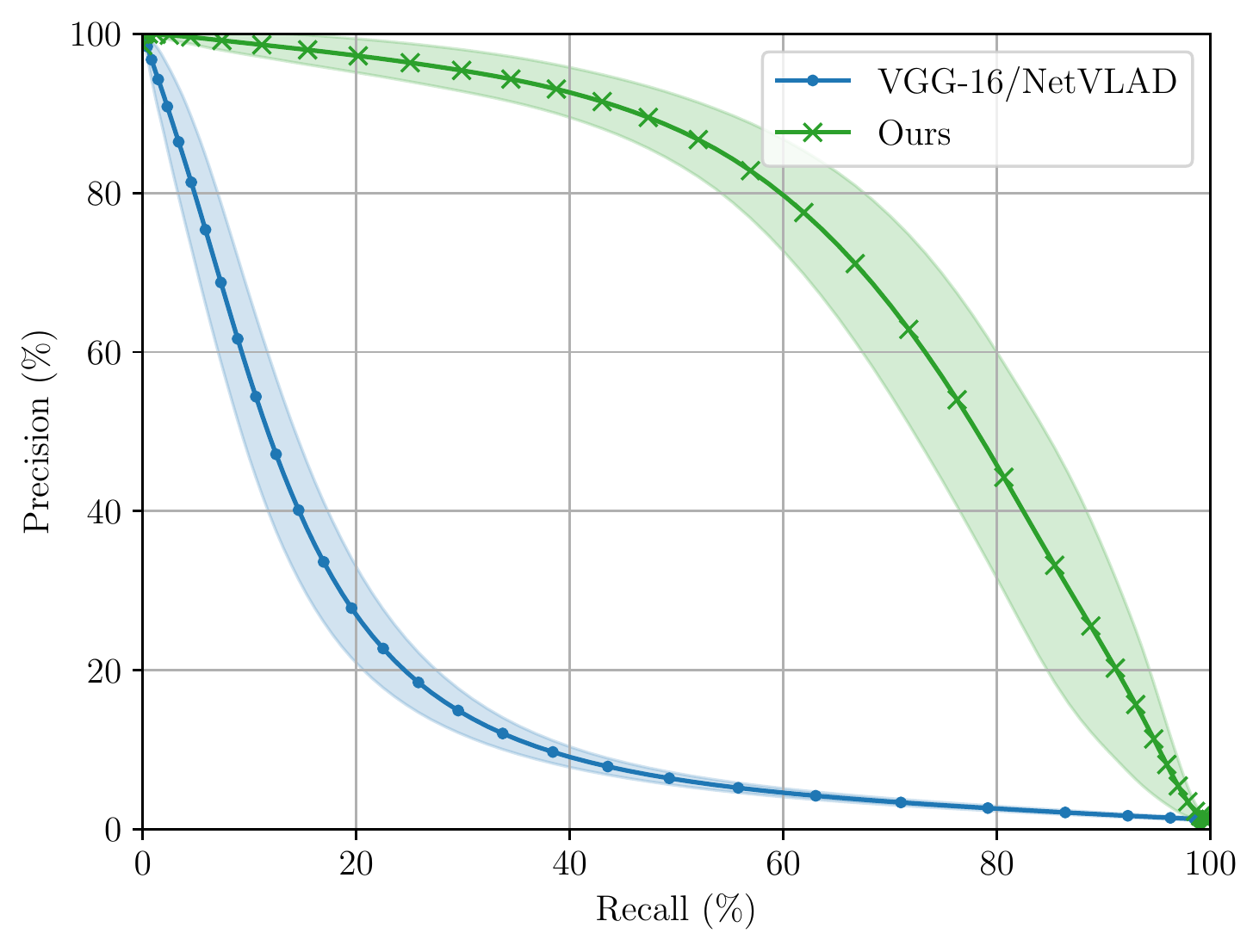}
    \caption{
    \gls{pr} curves representing localisation of all dataset trajectories against a static map when the input frames are randomly perturbed along the azimuth axis.
    In comparison to \cref{fig:rrcd_rr0_pr_curves}, the performance of \vanilla~degrades; with maximum $F_1$, $F_{0.5}$, and $F_2$ scores of $0.23 \pm 0.02$, $0.25 \pm 0.02$, and $0.30 \pm 0.02$ respectively.
    In contrast, our system -- having been designed to be quasi-rotationally invariant -- maintains approximately the same performance level; with maximum $F_1$, $F_{0.5}$, and $F_2$ scores of $0.69 \pm 0.04$, $0.70 \pm 0.04$, and $0.77 \pm 0.04$ respectively.
    }
    \label{fig:rrcd_rr1_pr_curves}
    \vspace{-0.5cm}
\end{figure}

\subsection{Performance metrics}
\label{sec:experimental:metrics}

In the ground truth $SE(2)$ database, all locations within a \SI{25}{\metre} radius of a ground truth location are considered true positives whereas those outside of a \SI{50}{\metre} radius are considered true negatives.

To evaluate \acrfull{pr}, we perform a ball search of the discretised metric space out to a varying embedding distance threshold.
While we show every third marker in the \gls{pr} curves to follow, we in fact typically evaluate \num{127} thresholds linearly spaced between the minimum and maximum values in an embedding distance matrix. 
As useful summaries of \gls{pr} performance, we pay heed to \gls{auc} as well as a swathe of F-scores, including $F_{1}$, $F_{2}$, and $F_{\beta}$ with $\beta = 0.5$~\cite{pino1999modern}.

As the eventual target application is part of a bigger system (c.f.~\cref{sec:introduction,sec:system}), we also defer to computational constraints and generate some systems-oriented metrics by varying the number of top-scoring candidates instead of a ball search.
To this end, we analyse the likely drop-out in localisation as the distance, $d$, the vehicle would have to travel on dead-reckoning (odometry) alone.
Additionally, a ``frames correctly localised'' quantity is calculated as the fraction of query frames along the trajectories to be localised at which at least one returned candidate was a true positive (bonafide localisation), with the assumption that a downstream verification process is capable of selecting that candidate (e.g. scan matching).

\section{Results}%
\label{sec:results}

This section presents instrumentation of the metrics discussed in~\cref{sec:experimental:metrics}.

After \num{500000} steps, the average ratio of embedding distance between positive and negative examples in the validation split was \SI{45.89}{\percent} (\radvlad) as compared to \SI{50.03}{\percent}, indicating better seperability in the learned metric space.
This corresponds to $F_1$ scores of \SI{90.49}{\percent} (\radvlad) as compared to \SI{89.98}{\percent} (\vanilla), $F_{0.5}$ scores of \SI{89.52}{\percent} (\radvlad) as compared to \SI{88.75}{\percent} (\vanilla), and $F_2$ scores of \SI{73.86}{\percent} (\radvlad) as compared to \SI{73.20}{\percent} (\vanilla).
We delay any decisive comparison of the utility and generalisability of the learned representations to the discussion below but what is worth noting here is that the architectures both perform better in the validation split than in the entire route, as the split was quite simple.

We then apply the learned metric space to encode an entire trajectory from the dataset (c.f.~\cref{sec:experimental:demarc}), including data from all splits (\textit{train}, \textit{valid}, and \textit{test}).
This encoded trajectory is used as a static map along which all other trajectories in the dataset are localised against.
We exclude the pair of trajectories which we use to train the network.
\cref{fig:rrcd_rr0_pr_curves} shows average \gls{pr} curves with one standard-deviation bounds.
The corresponding \gls{auc} are $0.75 \pm 0.06$ for \radvlad~as compared to $0.72 \pm 0.05$ for \vanilla.
This experiment serves to indicate that our proposed modifications result in measurable performance improvements over the baseline system.

We then better illustrate the rotational invariance of our proposed architecture by showing in~\cref{fig:test_rr0_pr_curves} average \gls{pr} curves when only data from the \textit{test} split is used for mapping and subsequent localisation.
This part of the environment was not seen by the network during training, and consists of a backwards traversal during which the vehicle is driving on the opposite side of the street (c.f.~\cref{fig:osm_se2}).
Here, the corresponding \gls{auc} is $0.52 \pm 0.04$ for \radvlad~as compared to $0.41 \pm 0.03$ for \vanilla.

Next, we use the static map built with all splits to localise incoming query frames which have been randomly perturbed along the azimuth axis, to probe the resilience to rotational disturbance.
\cref{fig:rrcd_rr1_pr_curves} shows degradation in the performance of \vanilla~while \radvlad~maintains the ability to recognise places.
Admittedly, we did not expect \vanilla~to perform well under these conditions, as we have performed no data augmentation which would account for this perturbation.
However, considering that \radvlad~was also trained on upright scans, this result vindicates the proposed architectural modifications.

\begin{figure}
    \centering
    \includegraphics[width=\columnwidth]{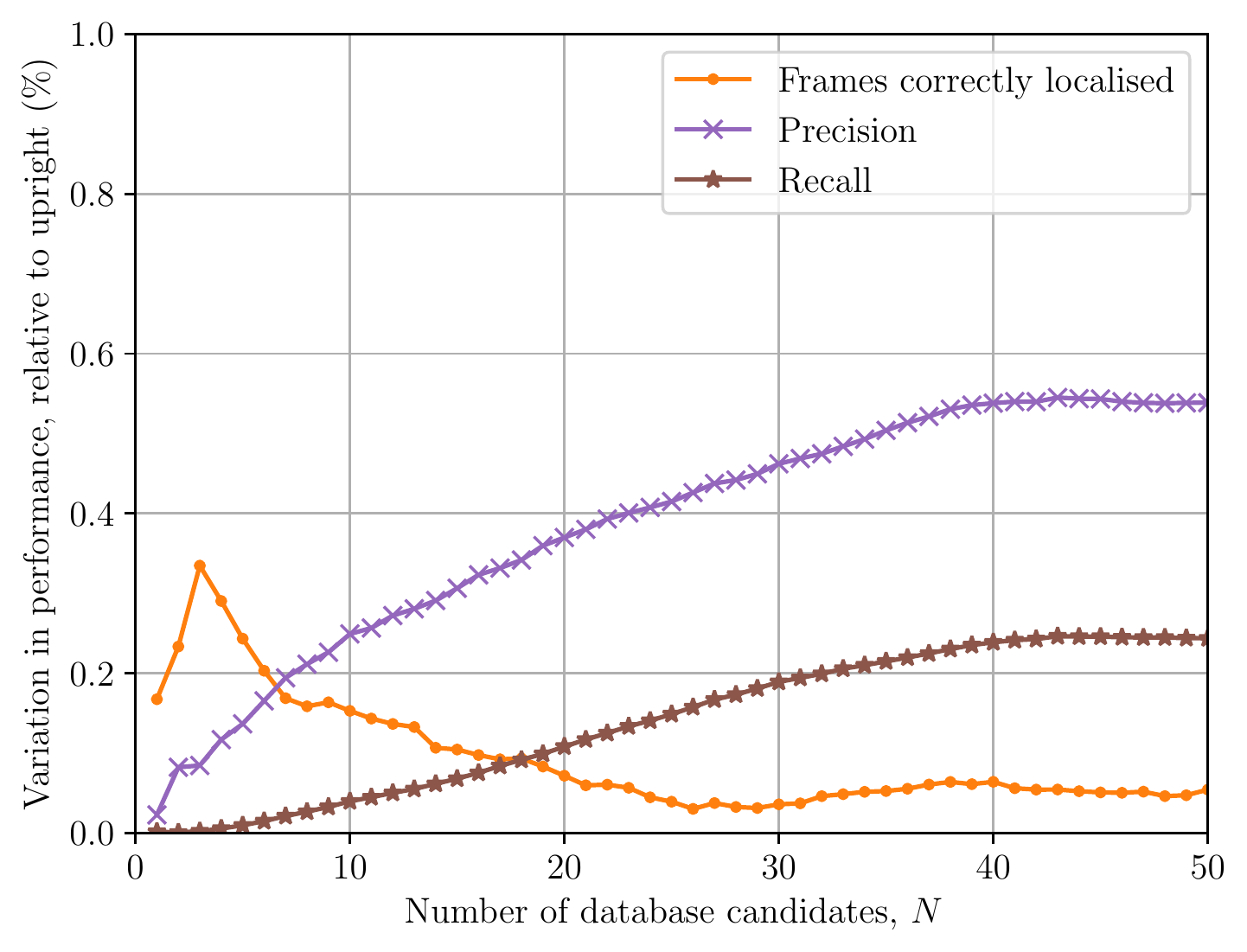}
    \caption{
    Variation in our system's performance when randomly rotated input frames are used for inference, as the fractional deviation from the ideal performance level when inference is performed upon unaltered, upright frames.
    The dependent axis is a percentage, not a fraction.
    The range of absolute values for the ideal performance against which this variation is measured is, for ``frames correctly localised'', (\SI{90.82}{\percent}, \SI{97.59}{\percent}).
    This means that, even when only considering the $1$-\gls{nn} in embedding space, we are able to localise \emph{correctly} \SI{90.82}{\percent} of the time.
    On the scale of the journeys represented here (\num{27} localised trajectories of about \SI{9}{\kilo\metre} each), this corresponds to approximately \SI{221}{\kilo\metre} of good localisations over a \SI{243}{\kilo\metre} drive.
    }
    \label{fig:top_db_rrcd}
    \vspace{-0.5cm}
\end{figure}

\cref{fig:top_db_rrcd} shows the performance of \radvlad~as a set of relative measures, where the upright condition is taken as the baseline signal, and the rotationally-perturbed condition is enumerated as fractional variation away from this ideal performance.
The independent variable shown in~\cref{fig:top_db_rrcd} is the number of database candidates which are closest in the embedding metric space which would have to be disambiguated by a downstream process (e.g. geometric verification in scan matching).
This is different to the threshold sweep used to generate the \gls{pr} curves (c.f.~\cref{fig:rrcd_rr0_pr_curves,fig:test_rr0_pr_curves,fig:rrcd_rr1_pr_curves}), which corresponds to a ball in the multidimensional space.
Each of these quantities is averaged over all \num{27} localisation trajectories against the same static map, as above.
The ranges of change in corresponding absolute values from $N = 1$ to $N = 50$ are (\SI{90.82}{\percent}, \SI{97.59}{\percent}) for frames correctly localised, (\SI{94.67}{\percent}, \SI{78.78}{\percent}) for precision, and (\SI{0.93}{\percent}, \SI{34.55}{\percent}) for recall, all for the upright condition.
We observe again that our system is extremely robust to rotational disturbances, where each of these systems metrics are within \SI{0.6}{\percent} of the ideal, upright condition.

\begin{figure}
    \centering
    \includegraphics[width=\columnwidth]{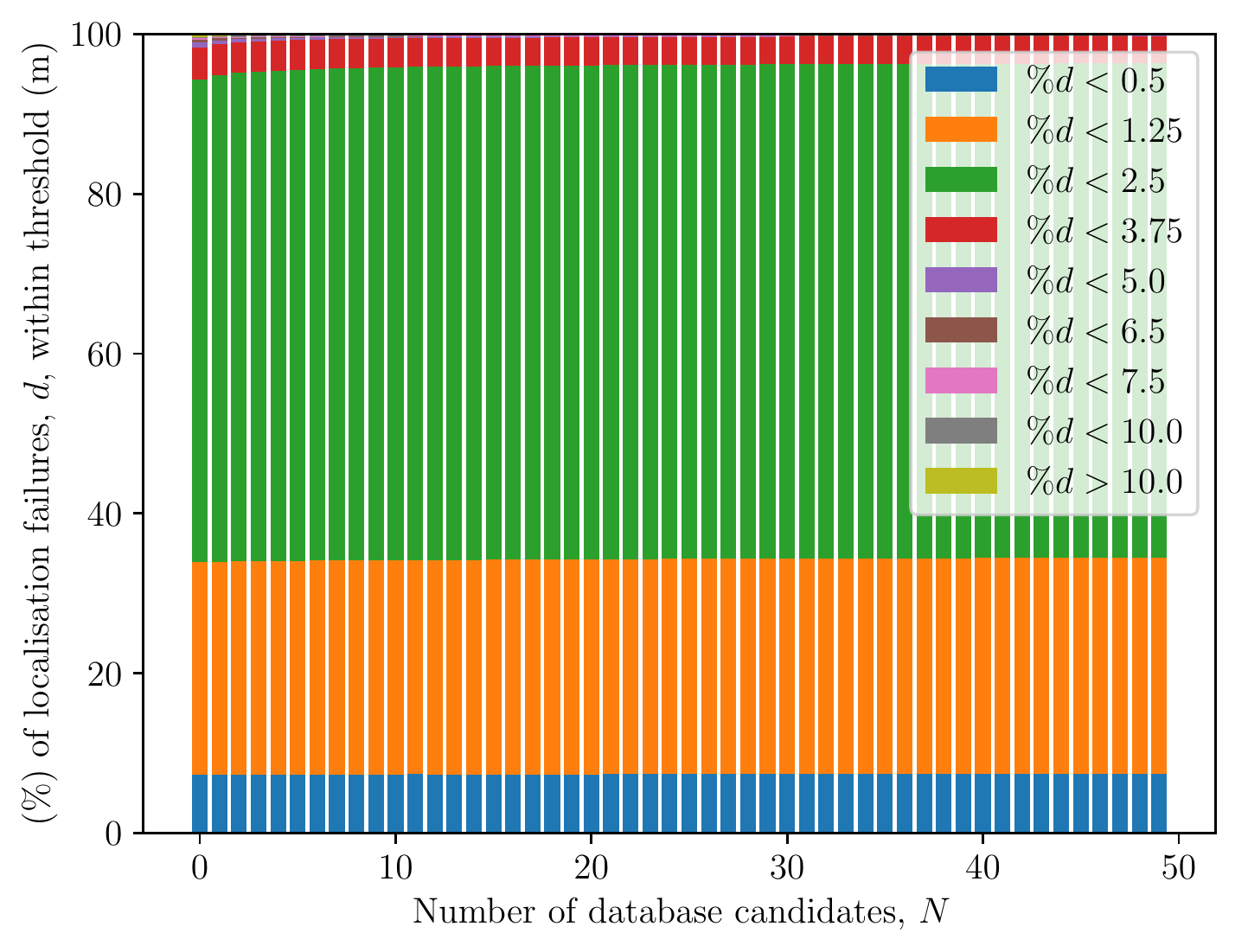}
    \caption{
    Percentage of localisation failure lengths within certain thresholds.
    A localisation failure is measured by the distance the vehicle would be required to travel on dead-reckoning (e.g. odometry) alone without a correct localisation.
    Note that all input frames to be localised have been randomly rotated, as in \cref{fig:top_db_rrcd}.
    These proportions are measured by combining the results of all localised trajectories.
    Using only the $1$-\gls{nn}, the proportion of failures limited to \SI{3.75}{\metre} is \SI{94.33}{\percent} and increases to \SI{96.34}{\percent} for $N = 50$.
    }
    \label{fig:ttime_netvlad-mod-pretrained_rrcd_rr1}
    \vspace{-0.3cm}
\end{figure}

Finally,~\cref{fig:ttime_netvlad-mod-pretrained_rrcd_rr1} shows the performance of our system as histograms of failure severity which are measured as the proportion of drop-outs in correct localisation results during which the vehicle moves a certain distance.
We observe that over \SI{90}{\percent} of the failures are limited to a driven distance of less than \SI{3.75}{\metre}, even when only the closest embedding in the metric space is taken as the localisation result.
As the number of candidates considered increases, this proportion tapers off as the worst failures are alleviated.
The worst measured failure for $N = 1$ is \SI{20.00}{\metre}, which decreases to \SI{9.33}{\metre} for $N = 50$.

\section{Conclusions and Future Work}%
\label{sec:conclusions}

We have presented a system for radar-only place recognition using a metric feature space which is learned in a rotationally-invariant fashion.
We described adjustments to off-the-shelf image recognition frameworks for better suitability to the native radar polar scan representation.
We demonstrated the efficacy of our approach on the largest radar-focused autonomous driving dataset collected to date, showing improved localisation capability compared to the na\"{i}ve application to the radar domain of competitive vision-based approaches, especially in the face of severe rotational disturbance.

In the future we plan to integrate the system presented in this paper with our mapping and localisation pipeline which is built atop of the scan-matching algorithm of~\cite{2018ICRA_cen,2019ICRA_cen} and to deploy the combined system in a teach-and-repeat autonomy scenario using the platform we have presented in~\cite{2019FSR_kyberd}, the conception of which was in large part concerned with deploying our \gls{fmcw} radar scanner.

\section*{Acknowledgments}

This project is supported by the Assuring Autonomy International Programme, a partnership between Lloyd’s Register Foundation and the University of York.
We would also like to thank our partners at Navtech Radar.
\cb{S}tefan S\u{a}ftescu is supported by UK's Engineering and Physical Sciences Research Council (EPSRC) through the Centre for Doctoral Training (CDT) in Autonomous Intelligent Machines and Systems (AIMS) Programme Grant EP/L015897/1.
Matthew Gadd is supported by Innovate UK under CAV2 -- Stream 1 CRD (DRIVEN).
Dan Barnes is supported by the UK EPSRC Doctoral Training Partnership.
Daniele De Martini and Paul Newman are supported by the UK EPSRC programme grant EP/M019918/1.

\bibliographystyle{IEEEtran}
\bibliography{biblio}

\begin{thebibliography}{10}
\providecommand{\url}[1]{#1}
\csname url@samestyle\endcsname
\providecommand{\newblock}{\relax}
\providecommand{\bibinfo}[2]{#2}
\providecommand{\BIBentrySTDinterwordspacing}{\spaceskip=0pt\relax}
\providecommand{\BIBentryALTinterwordstretchfactor}{4}
\providecommand{\BIBentryALTinterwordspacing}{\spaceskip=\fontdimen2\font plus
\BIBentryALTinterwordstretchfactor\fontdimen3\font minus
  \fontdimen4\font\relax}
\providecommand{\BIBforeignlanguage}[2]{{%
\expandafter\ifx\csname l@#1\endcsname\relax
\typeout{** WARNING: IEEEtran.bst: No hyphenation pattern has been}%
\typeout{** loaded for the language `#1'. Using the pattern for}%
\typeout{** the default language instead.}%
\else
\language=\csname l@#1\endcsname
\fi
#2}}
\providecommand{\BIBdecl}{\relax}
\BIBdecl

\bibitem{OpenStreetMap}
{OpenStreetMap contributors}, ``{Planet dump retrieved from
  https://planet.osm.org },'' \url{ https://www.openstreetmap.org }, 2017.

\bibitem{haklay2008openstreetmap}
M.~Haklay and P.~Weber, ``{OpenStreetMap: User-generated street maps},''
  \emph{IEEE Pervasive Computing}, vol.~7, no.~4, pp. 12--18, 2008.

\bibitem{2018ICRA_cen}
S.~H. Cen and P.~Newman, ``Precise ego-motion estimation with millimeter-wave
  radar under diverse and challenging conditions,'' \emph{Proceedings of the
  2018 IEEE International Conference on Robotics and Automation}, 2018.

\bibitem{2019ICRA_cen}
S.~Cen and P.~Newman, ``Radar-only ego-motion estimation in difficult settings
  via graph matching,'' in \emph{Proceedings of the IEEE International
  Conference on Robotics and Automation (ICRA), Montreal, Canada}, 2019.

\bibitem{2019ITSC_aldera}
R.~Aldera, D.~De~Martini, M.~Gadd, and P.~Newman, ``{What Could Go Wrong?
  Introspective Radar Odometry in Challenging Environments},'' in \emph{{IEEE
  Intelligent Transportation Systems (ITSC) Conference}}, Auckland, New
  Zealand, October 2019.

\bibitem{2019ICRA_aldera}
R.~\vspace{0mm}Aldera, D.~De~Martini, M.~Gadd, and P.~Newman, ``Fast radar
  motion estimation with a learnt focus of attention using weak supervision,''
  in \emph{Proceedings of the IEEE International Conference on Robotics and
  Automation (ICRA), Montreal, Canada}, 2019.

\bibitem{barnes2019masking}
\BIBentryALTinterwordspacing
D.~Barnes, R.~Weston, and I.~Posner, ``Masking by moving: Learning
  distraction-free radar odometry from pose information,'' \emph{arXiv preprint
  arXiv: 1909.03752}, 2019. [Online]. Available:
  \url{https://arxiv.org/pdf/1909.03752}
\BIBentrySTDinterwordspacing

\bibitem{cummins2008fab}
M.~Cummins and P.~Newman, ``{FAB-MAP: Probabilistic localization and mapping in
  the space of appearance},'' \emph{The International Journal of Robotics
  Research}, vol.~27, no.~6, pp. 647--665, 2008.

\bibitem{milford2012seqslam}
M.~J. Milford and G.~F. Wyeth, ``Seqslam: Visual route-based navigation for
  sunny summer days and stormy winter nights,'' in \emph{2012 IEEE
  International Conference on Robotics and Automation}.\hskip 1em plus 0.5em
  minus 0.4em\relax IEEE, 2012, pp. 1643--1649.

\bibitem{khaliq2018holistic}
A.~Khaliq, S.~Ehsan, M.~Milford, and K.~McDonald-Maier, ``A holistic visual
  place recognition approach using lightweight cnns for severe viewpoint and
  appearance changes,'' \emph{arXiv preprint arXiv:1811.03032}, 2018.

\bibitem{dube2017segmatch}
R.~Dub{\'e}, D.~Dugas, E.~Stumm, J.~Nieto, R.~Siegwart, and C.~Cadena,
  ``Segmatch: Segment based place recognition in 3d point clouds,'' in
  \emph{2017 IEEE International Conference on Robotics and Automation
  (ICRA)}.\hskip 1em plus 0.5em minus 0.4em\relax IEEE, 2017, pp. 5266--5272.

\bibitem{gawel2018x}
A.~Gawel, C.~Del~Don, R.~Siegwart, J.~Nieto, and C.~Cadena, ``X-view:
  Graph-based semantic multi-view localization,'' \emph{IEEE Robotics and
  Automation Letters}, vol.~3, no.~3, pp. 1687--1694, 2018.

\bibitem{angelina2018pointnetvlad}
M.~Angelina~Uy and G.~Hee~Lee, ``Pointnetvlad: Deep point cloud based retrieval
  for large-scale place recognition,'' in \emph{Proceedings of the IEEE
  Conference on Computer Vision and Pattern Recognition}, 2018, pp. 4470--4479.

\bibitem{guo2019local}
J.~Guo, P.~V. Borges, C.~Park, and A.~Gawel, ``Local descriptor for robust
  place recognition using lidar intensity,'' \emph{IEEE Robotics and Automation
  Letters}, vol.~4, no.~2, pp. 1470--1477, 2019.

\bibitem{reina2015radar}
G.~Reina, D.~Johnson, and J.~Underwood, ``Radar sensing for intelligent
  vehicles in urban environments,'' \emph{Sensors}, vol.~15, no.~6, pp.
  14\,661--14\,678, 2015.

\bibitem{mielle2019comparative}
M.~Mielle and a.~L. A.~J. Magnusson, Martin, ``{A comparative analysis of radar
  and lidar sensing for localization and mapping},'' in \emph{Proceedings of
  the European Conference on Mobile Robotics (ECMR)}, 2019.

\bibitem{takeuchi2015localization}
E.~Takeuchi, A.~Elfes, and J.~Roberts, ``Localization and place recognition
  using an ultra-wide band (uwb) radar,'' in \emph{Field and service
  robotics}.\hskip 1em plus 0.5em minus 0.4em\relax Springer, 2015, pp.
  275--288.

\bibitem{figueroa2016comparison}
A.~Figueroa, B.~Al-Qudsi, N.~Joram, and F.~Ellinger, ``{Comparison of two-way
  ranging with FMCW and UWB radar systems},'' in \emph{2016 13th Workshop on
  Positioning, Navigation and Communications (WPNC)}.\hskip 1em plus 0.5em
  minus 0.4em\relax IEEE, 2016, pp. 1--6.

\bibitem{cadena2016past}
C.~Cadena, L.~Carlone, H.~Carrillo, Y.~Latif, D.~Scaramuzza, J.~Neira, I.~Reid,
  and J.~J. Leonard, ``Past, present, and future of simultaneous localization
  and mapping: Toward the robust-perception age,'' \emph{IEEE Transactions on
  robotics}, vol.~32, no.~6, pp. 1309--1332, 2016.

\bibitem{bentley1975multidimensional}
J.~L. Bentley, ``Multidimensional binary search trees used for associative
  searching,'' \emph{Communications of the ACM}, vol.~18, no.~9, pp. 509--517,
  1975.

\bibitem{malkov16}
Y.~A. Malkov and D.~A. Yashunin, ``Efficient and robust approximate nearest
  neighbor search using hierarchical navigable small world graphs,''
  \emph{CoRR}, vol. abs/1603.09320, 2016.

\bibitem{wieschollek_2016_CVPR}
P.~Wieschollek, O.~Wang, A.~Sorkine-Hornung, and H.~P.~A. Lensch, ``Efficient
  large-scale approximate nearest neighbor search on the gpu,'' in \emph{The
  IEEE Conference on Computer Vision and Pattern Recognition (CVPR)}, June
  2016.

\bibitem{arandjelovic2016netvlad}
R.~Arandjelovic, P.~Gronat, A.~Torii, T.~Pajdla, and J.~Sivic, ``Netvlad: Cnn
  architecture for weakly supervised place recognition,'' in \emph{Proceedings
  of the IEEE conference on computer vision and pattern recognition}, 2016, pp.
  5297--5307.

\bibitem{simonyan2014very}
K.~Simonyan and A.~Zisserman, ``Very deep convolutional networks for
  large-scale image recognition,'' \emph{arXiv preprint arXiv:1409.1556}, 2014.

\bibitem{cieslewski2018data}
T.~Cieslewski, S.~Choudhary, and D.~Scaramuzza, ``Data-efficient decentralized
  visual slam,'' in \emph{2018 IEEE International Conference on Robotics and
  Automation (ICRA)}.\hskip 1em plus 0.5em minus 0.4em\relax IEEE, 2018, pp.
  2466--2473.

\bibitem{wang2018omnidirectional}
T.-H. Wang, H.-J. Huang, J.-T. Lin, C.-W. Hu, K.-H. Zeng, and M.~Sun,
  ``Omnidirectional cnn for visual place recognition and navigation,'' in
  \emph{2018 IEEE International Conference on Robotics and Automation
  (ICRA)}.\hskip 1em plus 0.5em minus 0.4em\relax IEEE, 2018, pp. 2341--2348.

\bibitem{zhang2019making}
R.~Zhang, ``Making convolutional networks shift-invariant again,'' \emph{arXiv
  preprint arXiv:1904.11486}, 2019.

\bibitem{schroff2015facenet}
F.~Schroff, D.~Kalenichenko, and J.~Philbin, ``Facenet: A unified embedding for
  face recognition and clustering,'' in \emph{Proceedings of the IEEE
  conference on computer vision and pattern recognition}, 2015, pp. 815--823.

\bibitem{bengio2012practical}
Y.~Bengio, ``Practical recommendations for gradient-based training of deep
  architectures,'' in \emph{Neural networks: Tricks of the trade}.\hskip 1em
  plus 0.5em minus 0.4em\relax Springer, 2012, pp. 437--478.

\bibitem{goodfellow2016deep}
I.~Goodfellow, Y.~Bengio, and A.~Courville, \emph{Deep learning}.\hskip 1em
  plus 0.5em minus 0.4em\relax MIT press, 2016.

\bibitem{RobotCarDatasetIJRR}
W.~Maddern, G.~Pascoe, C.~Linegar, and P.~Newman, ``{1 Year, 1000km: The Oxford
  RobotCar Dataset},'' \emph{The International Journal of Robotics Research
  (IJRR)}, vol.~36, no.~1, pp. 3--15, 2017.

\bibitem{OxfordRadarRobotCarDataset}
\BIBentryALTinterwordspacing
D.~Barnes, M.~Gadd, P.~Murcutt, P.~Newman, and I.~Posner, ``{The Oxford Radar
  RobotCar Dataset: A Radar Extension to the Oxford RobotCar Dataset},''
  \emph{arXiv preprint arXiv: 1909.01300}, 2019. [Online]. Available:
  \url{https://arxiv.org/pdf/1909.01300}
\BIBentrySTDinterwordspacing

\bibitem{barnes2018driven}
D.~Barnes, W.~Maddern, G.~Pascoe, and I.~Posner, ``{Driven to distraction:
  Self-supervised distractor learning for robust monocular visual odometry in
  urban environments},'' in \emph{2018 IEEE International Conference on
  Robotics and Automation (ICRA)}.\hskip 1em plus 0.5em minus 0.4em\relax IEEE,
  2018, pp. 1894--1900.

\bibitem{pino1999modern}
J.~A. Pino, ``Modern information retrieval. ricardo baeza-yates y berthier
  ribeiro-neto addison wesley hariow, england, 1999,'' 1999.

\bibitem{2019FSR_kyberd}
S.~Kyberd, J.~Attias, P.~Get, P.~Murcutt, C.~Prahacs, M.~Towlson, S.~Venn,
  A.~Vasconcelos, M.~Gadd, D.~De~Martini, and P.~Newman, ``{The Hulk: Design
  and Development of a Weather-proof Vehicle for Long-term Autonomy in Outdoor
  Environments},'' in \emph{{International Conference on Field and Service
  Robotics (FSR)}}, Tokyo, Japan, August 2019.

\end{thebibliography}

\end{document}